\documentclass[runningheads]{llncs}
\usepackage{makeidx}
\usepackage{graphicx}
\usepackage{amsmath,amssymb} % define this before the line numbering.
\usepackage{color}
\usepackage{booktabs} % add line space in tables
\usepackage{siunitx} % add mirometer
\usepackage{hyperref} % use autoref
\usepackage{subcaption} % image sub stuff
\usepackage{bbm} % math symbols
\usepackage{placeins} % Float barriers
\usepackage{makecell}% line breaks in tables
\usepackage{float} % force position
\usepackage[utf8]{inputenc} % üäö
\usepackage{xcolor}

\newcolumntype{L}[1]{>{\raggedright\let\newline\\\arraybackslash\hspace{0pt}}m{#1}}

\newcommand{\br}{\vspace{1.5mm}\noindent}

\newcommand{\pic}[2]{
	\begin{figure}[htb] 
		\centering
		\includegraphics[width=\linewidth]{#1}
		\caption{#2}
		\label{fig:#1}
	\end{figure}
}

\newcommand{\incep}{Inception-ResNet-3D }

\begin{document}
	\pagestyle{headings}
	\mainmatter

	\title{2D and 3D Segmentation of uncertain local collagen fiber orientations in SHG microscopy}

	\titlerunning{3D Segmentation of uncertain orientations of collagen fibers}
	\authorrunning{L. Schmarje et al.}
	
	\author{Lars Schmarje\inst{1} \and
		Claudius Zelenka\inst{1}  \and
		Ulf Geisen\inst{2} \and
		Claus-C. Glüer\inst{2} \and
		Reinhard Koch\inst{1}
	}
	
	\institute{Multimedia Information Processing Group, Kiel University, Germany \email{\{las,cze,rk\}@informatik.uni-kiel.de}  \and
		Molecular Imaging North Competence Center, Kiel University, Germany\\
		\email{\{ulf.geisen,glueer\}@rad.uni-kiel.de}}

	\maketitle

\begin{abstract}
	
	Collagen fiber orientations in bones, visible with Second Harmonic Generation (SHG) microscopy, represent the inner structure and its alteration due to influences like cancer. 
	While analyses of these orientations are valuable for medical research, it is not feasible to analyze the needed large amounts of local orientations manually.
	Since we have uncertain borders for these local orientations only rough regions can be segmented instead of a pixel-wise segmentation. 
	We analyze the effect of these uncertain borders on human performance by a user study.
	Furthermore, we compare a variety of 2D and 3D methods such as classical approaches like Fourier analysis with state-of-the-art deep neural networks for the classification of local fiber orientations.
	We present a general way to use pretrained 2D weights in 3D neural networks, such as Inception-ResNet-3D a 3D extension of Inception-ResNet-v2.
	In a 10 fold cross-validation our two stage segmentation based on Inception-ResNet-3D and transferred 2D ImageNet weights achieves a human comparable accuracy.
		
	
\end{abstract}

\section{Introduction}

In a variety of medical issues and research activities, computed tomography (CT) scans are used for bone examinations.
However, most CT scans only have a resolution in the millimeter range.
Special CT procedures allow resolutions of a few \si{\um}  \cite{ct-imaging,ct-spatial}.
For this reason single collagen fiber bundles of about 2-3 \si{\um} \cite{pork} can not be detected well in CT scans.
The structure and orientation of these bundles allow us to make conclusions about changes in the bone (e.g. by growth or disease) \cite{pork}.
These characteristics of the inner bone structure are valuable for research in the fields of age determination, disease detection and cancer research.

\br
Second harmonic generation (SHG) microscopy can visualize these structures of collagen due to its higher resolution.
This methods allows us to generate large amounts of dense 3D scans of collagen fibers in bones.
It is time-consuming to create statistics of fiber bundles orientations or to mark regions of interest by experts.
Moreover, manual annotations for large datasets are not feasible due to time constraints, budget and subjective biases.
These biases are results of the uncertain borders in local fiber orientations.
Therefore, an automatic analysis of fiber bundles orientation in large amounts of SHG data would benefit a variety of medical research activities.
Large scale studies are not practical without automatic analysis.

\br
The disease osteogenesis imperfecta (OI), also known as brittle bone disease, changes the orientation of fiber bundles in the bones of affected people and animals\cite{pork}.
Hence SHG data of healthy and diseased mice is predestined for the evaluation of new methods.

\begin{figure}[tb] 
	\centering
	\begin{subfigure}{.3\textwidth}
		\includegraphics[width=\linewidth]{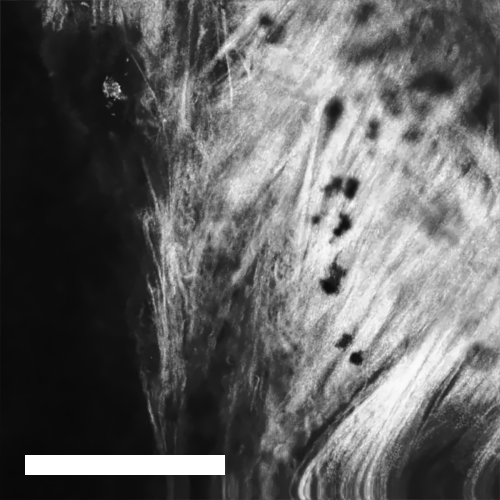}
		\caption{SHG image}
	\end{subfigure}
	\begin{subfigure}{.3\textwidth}
		\includegraphics[width=\linewidth]{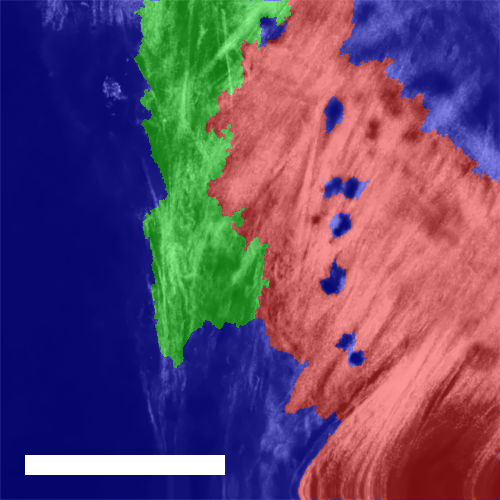}
		\caption{Pixelwise GT}
	\end{subfigure}
	\begin{subfigure}{.3\textwidth}
		\includegraphics[width=\linewidth]{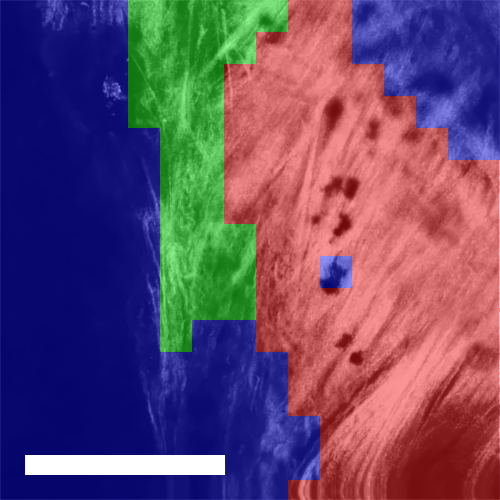}
		\caption{Rough GT}
	\end{subfigure}
	\caption{
		Example of a SHG input image and the corresponding ground-truth -
		Note that while (b) is a pixel-wise ground-truth annotation we are in a setting with uncertain borders.
		Hence we should try to recreate a rough segmentation as shown in (c).
		The white scale bar depicts a size of 100 \si{\um}.
		Color code: Similar orientation - red, Dissimilar orientation - green, Not of interest - blue
	}
	\label{fig:input-gt}
\end{figure}

\br
Consequently, we developed different automatic algorithms for fiber bundle orientation analysis.
We considered the orientation of fiber bundles in a local region as a classification problem.
We focused ourselves on three classes: Bundles with similar (S) and dissimilar (D) orientations and everything else or not of interest (N) (e.g. noise, background).
\autoref{fig:input-gt} shows an example input and the corresponding ground-truth segmentation images.

\br
The classification of a local region can also be defined as a rough semantic segmentation of the entire image.
Since borders of local orientations are not well defined and can be described as highly uncertain and fluent, the goal is not to create pixel-wise segmentation.
Therefore, we are interested in rough localization of these regions and their classification.
Due to large regions which are not of interest (N) a large class imbalance in favor of this class exists and must be addressed.
We analyzed the effect of uncertain borders on human performance in the task of rough fiber orientation segmentation by a user study.

\br
We investigated classical approaches like Fourier analysis and state-of-the-art methods like deep neural networks.
Instead of reporting only the best results we present a complete overview and comparison for future research in rough semantic segmentation.
Most of our neural networks use a state-of-the-art backbone and domain specific adaptions.
We aimed to change as little as possible in the original backbones to allow interchangeability with other backbones in the future.
We show how a state-of-the-art 2D backbone can be used in 3D rough semantic segmentation.
We call this network Inception-ResNet-3D.
Especially we present a way to transfer pretrained 2D weights into the 3D case.
The code is publicly available for reproducibility. \footnote[1]{https://github.com/Emprime/uncertain-fiber-segmentation}

\br
To sum up, the main contributions of our work are:
\begin{itemize}
	\item We report a systematical comparison of algorithms for classification and segmentation in 2D and 3D with uncertain borders.
	We use a novel dense 3D SHG dataset with more than 4500 slices for method development and testing.
	Human performance to classify uncertain collagen fiber orientations on this dataset is also reported.
	\item We show a general way to convert weights from the 2D into 3D.
	\item Our two stage approach with \incep and transferred 2D weights achieves a performance comparable to humans for uncertain collagen fiber segmentation in a 10 fold cross-validation.
\end{itemize}

\section{Related Work}

Currently neural networks are state-of-the-art in the field of image data classification (e.g. ImageNet \cite{imagenet}).
A variety of neural networks have emerged over the years \cite{alexnet,vgg16,resnet,densenet,inception,incepresv2}.
These networks started with a simple architecture (e.g. VGG-16 \cite{vgg16}).
They integrated new structure elements like residual \cite{resnet} and inception blocks \cite{inception} as they were developed and proved their superior performance.
This development led to an increase of the top1 accuracy on the ImageNet test set from 71.3\% with VGG-16 to 80.3\% with Inception-ResNet-v2.
Parallel the depth and thereby the complexity increased from 23 to 572 layers \footnote[2]{Values are based on the reference implementation in \href{https://keras.io/applications/}{Keras}}.

\br
Semantic segmentation gives a classification for every pixel in an image and is an extension of a classification problem.
Shelhamer et al. \cite{fcn} first proposed to use fully convolutional networks to solve semantic segmentation.
U-Net \cite{unet} is a network for semantic segmentation which was designed for medical images.
Often semantic segmentation networks consist of a down- and a upsampling part \cite{fcn,unet}.

\br
However, the current state-of-the-art approaches for image classification and semantic segmentation have two major drawbacks in the context of uncertain local fiber orientation classification.
We have 3D data and a high uncertainty for the borders.
Most research focuses on 2D data while Zhou et al. \cite{2d-3d-net} showed that it is beneficial to use the 3D information for organ segmentation.
Networks like PointNet \cite{pointnet} can classify 3D point clouds yet they do not consider dense 3D input as we have.
The network 3D-U-Net \cite{3d-unet} represents an expansion of U-Net to 3D data.
It is typically used to segment 3D objects like organs \cite{3d-unet}.
This fixes the first drawback while the second one remains.
Objects with uncertain borders like our fiber orientations are not well represented.

\br
While 3D extensions of Inception-ResNet-v2 have been presented in \cite{incep3d,incep3D-2} the usage of 2D pretraining is not so widely used.
Parallel to our research Shan et al. proposed a 2D weight transfer strategy to 3D \cite{weightTransfer} which is most similar to ours (see \autoref{subsec:weight}).

\br
Collagen structures in SHG images have been analyzed in several publications \cite{shg-tissue,ancient,thg,pork,horse,ft-shg,ft-shg-3d}.
They were analyzed in tissue \cite{shg-tissue} and bones \cite{thg,ancient}.
Rao et al. \cite{ft-shg} presented how Fourier analysis can be used to investigate the orientation of collagen fibers.
The Fourier analysis was extended from small regions to the whole scan in \cite{pork,horse,ft-shg-3d}.
The analysis classified small image parts as anisotropic, isotropic and dark.
These classifications where used to calculate the distributions of classes over an image.
In \cite{horse} these distributions where used to detect injured tendons.
Moreover, Ambekar et al. \cite{pork} showed the change of distribution due to aging can be used to determine the age of pigs.
Lau et al. \cite{ft-shg-3d} used the 3D information of SHG data and could show an increase in performance.

\br
Nevertheless, their analysis is based on only few ($<$100) images.
An analysis on larger amounts of data is not known.
The data shown in the papers seems to be of overall of a good quality.
Artifacts, noise and blurring and impact of performance was not reported.

\br
Liang et al. \cite{shg-nn} state to be the first to analyze SHG images with neural networks.
They estimated the elastic properties of collagenous tissue.
A classification or segmentation of fibers were not part of their investigation.

\br
To our knowledge, we are the first who use neural networks to automatically classify and segment local collagen fiber orientation in large amounts of 2D or 3D data.
In contrast to previous neural network literature we use 3D data and adapt our networks to uncertain borders.
In comparison to earlier fiber analyses we utilize neural networks to process large amount of mixed quality data.

\section{User study}
\label{sec:pre}

While we knew that we operated in a context with uncertain borders we did not know how this would impact performance.
Therefore, we investigated this issue by a random sample user study.
Our goal was to examine how well humans can classify and segment local fiber orientations.
We compared 15 different people with each other (interpersonal) and 5 results of the same person over time (intrapersonal).

\br
The participants were given two tasks.
The first task was to chose one annotation out of 5 given example annotations for one image.
This task was repeated for 10 different images.
The second task was to create an annotation for 24 images.

\br
For the first task we calculated the Pearson correlation coefficient between the annotation selections of all participants (interpersonal) or over time (intrapersonal).
This leads to a mean absolute coefficient of 0.44 with a standard deviation of 0.26 for the intrapersonal comparison.
The interpersonal comparison results in a mean absolute coefficient of 0.24 with a standard deviation of 0.2.

\br
For the second task we calculated the accuracies of the created annotations with the ground-truth (see \autoref{subsec:metric} for the metric definition).
The intrapersonal comparison reached a mean accuracy of 78.29\% with a standard deviation of 2.40\% over all 24 images.
The interpersonal comparison resulted in a mean accuracy of 58.83\% with a standard deviation of 7.44\%.

\br
All in all we see that it is more difficult for different people to select or create consistent annotations than for one person over time.
However, even for a person over time the selection and creation is not perfect. 
We can state that humans achieve only about 78\% accuracy consistency with themselves.
If we train and evaluate a neural network on human created ground-truth with this consistency rate we can not expect that an algorithm performs significantly better.

\section{Methods}

All methods use the same datasets although the 2D methods ignore the inherent three-dimensional information.
Therefore, 2D data will be referred to as scan slice or image and 3D data as scan.
For all methods we investigated a variety of hyperparameters such as batchsize, backbones and loss variations.
We will mention in the method description only important hyperparameter selections and specialties.
For further details see the supplementary materials.

\subsection{Weight transfer 2D to 3D}
\label{subsec:weight}

We want to utilize the pretrained ImageNet weights in our 3D Networks and thus we have to transfer the 2D kernel weights into 3D kernel weights.
Technically this is a function $I: \mathbb{R}^{w \times h \times c} \rightarrow \mathbb{R}^{w \times h \times d \times c}$ that transforms a 3D (width, height, channels) matrix $M_1$ into a 4D (width, height, depth, channels) matrix $M_2$ with $w,h,d,c \in \mathbb{N}$ .

\br
We investigated two methods for the weight transformation.
We denote the set $\{1,..,N\}$ by $[n]$ .
The first approach is to divide $M_1$ by $d$ and stack them $d$ times to create $M_2$ for a given depth $d \in \mathbb{N}$:

\begin{equation}
\label{eg:transfer1}
M_{2 (i,j,k,l)} = M_{1 (i,j,l)} / d
\mbox{ for all } i \in [w],
j \in [h],
k \in [d],
\mbox{ and } l \in [c].
\end{equation}
\\
The second approach is to insert the 3D matrix into the 4D matrix and fill the rest up with zeros.
Shan et al. proposed in \cite{weightTransfer} a similar method.
For given odd depth $d \in \mathbb{N}$ and center element $\hat{c} = \frac{(d-1)}{2} +1$ this is defined as

\begin{equation}
\label{eg:transfer2}
M_{2 (i,j,k,l)} = 
	\begin{cases}
	M_{1 (i,j,l)} \mbox{ for } k = \hat{c}\\ 
	0 \mbox{ \hspace{1.3cm} for } k \in [d] \setminus \{\hat{c}\}
	\end{cases}
\end{equation}
and all $i \in [w], j \in [h], k \in [d], m \in [c]$.

\br
Henceforth, we refer to theses transformations if we talk about 2D weights in a 3D context or transferred weights.
See \autoref{subsec:3d-methods} for further details on the selected transfer strategies.
We use this weight transformations in the network Inception-Resnet-3D a 3D extension of Inception-ResNet-v2 \cite{incepresv2}.
In general the architecture is the same but with 3D layers as done before by \cite{incep3d,incep3D-2}.
In the case of asymmetric input data we introduce asymmetric strides in the downsamplings in the stem block.
These asymmetric strides create symmetric input for deeper layers.

\subsection{Weighted Focal Loss}

As mentioned before we have to address the issue of class imbalance in our training data and chose to investigate different loss variations.
Lin et al. defined the novel loss function Focal Loss in \cite{focalloss} which should automatically balance the contribution of classes in skewed cases.
As mentioned in \cite{focalloss} the loss $L: [0,1]^n \times \{0,1\}^n \rightarrow [0,1]$ can be extended to the non binary case as shown in \autoref{eq:weighted-focal-loss} with $n \in \mathbb{N}$ being the number of classes and $\gamma$ the Focal Loss parameter.
Furthermore, we integrated weights $w \in \mathbb{R}^n$ for every prediction $ \hat{y} \in [0,1] ^n $ and ground-truth $ y \in \{ 0,1 \}^n $.
In our case $n=3$ and the values of $y$ and $\hat{y}$ correspond to these three classes.

\begin{equation}
\label{eq:weighted-focal-loss}
L(\hat{y},y) = - \sum_{i \in [n]} w_i (1-\hat{y_i})^\gamma y_i log(\hat{y_i})
\end{equation}

\br
Keep in mind that \autoref{eq:weighted-focal-loss} is equal to cross entropy if $\gamma = 0$ and $w = \{1\}^3$.

\subsection{2D methods}
\label{subsec:2d-methods}

\subsubsection{Fourier analysis}

As a reference we reimplemented a classification based on the Fourier analysis in small image patches \cite{ft-shg}.
We used thresholds like in \cite{ft-shg} to discriminate different classes.

\begin{figure}[tb] 
	\centering
	\begin{subfigure}{.3\textwidth}
		\includegraphics[width=\linewidth]{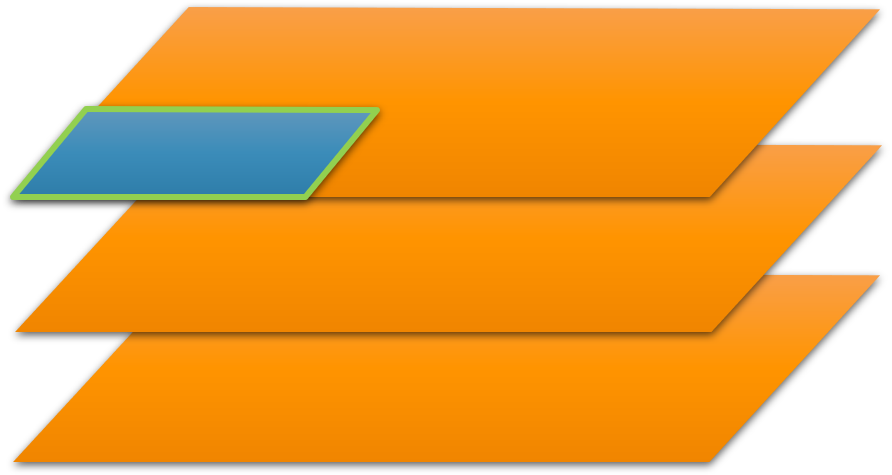}
		\caption{2D classification}
		\label{fig:graphical-a}
	\end{subfigure}
	\begin{subfigure}{.3\textwidth}
		\includegraphics[width=\linewidth]{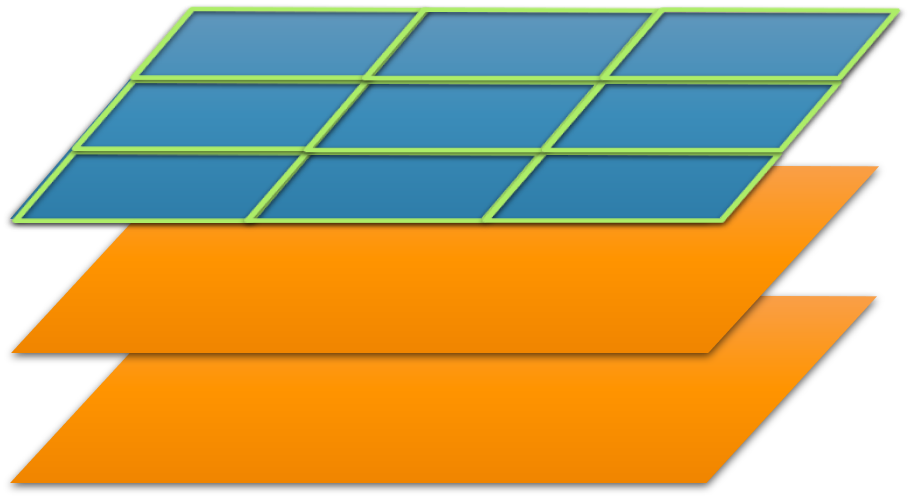}
		\caption{2D semantic segmentation}
		\label{fig:graphical-b}
	\end{subfigure}
	\begin{subfigure}{.3\textwidth}
		\includegraphics[width=\linewidth]{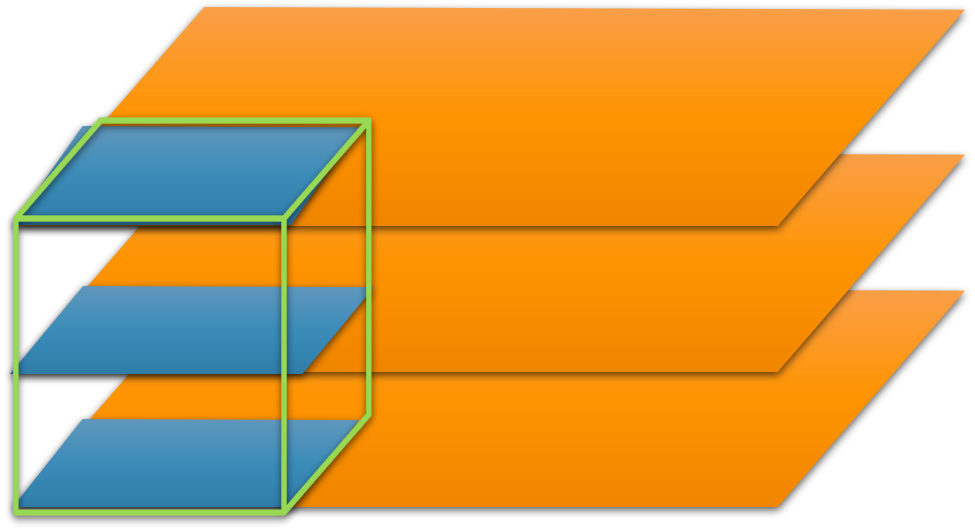}
		\caption{3D combination of 2D classifications}
		\label{fig:graphical-c}
	\end{subfigure}
	\begin{subfigure}{.3\textwidth}
		\includegraphics[width=\linewidth]{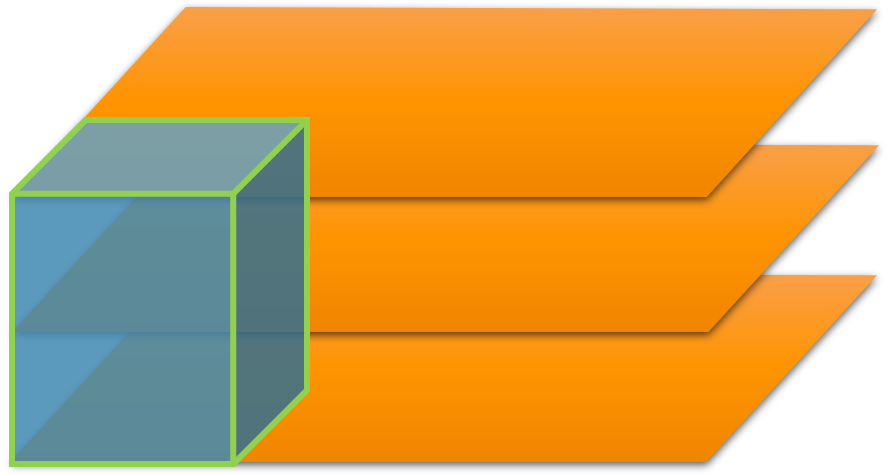}
		\caption{3D classification}
		\label{fig:graphical-d}
	\end{subfigure}
	\begin{subfigure}{.3\textwidth}
		\includegraphics[width=\linewidth]{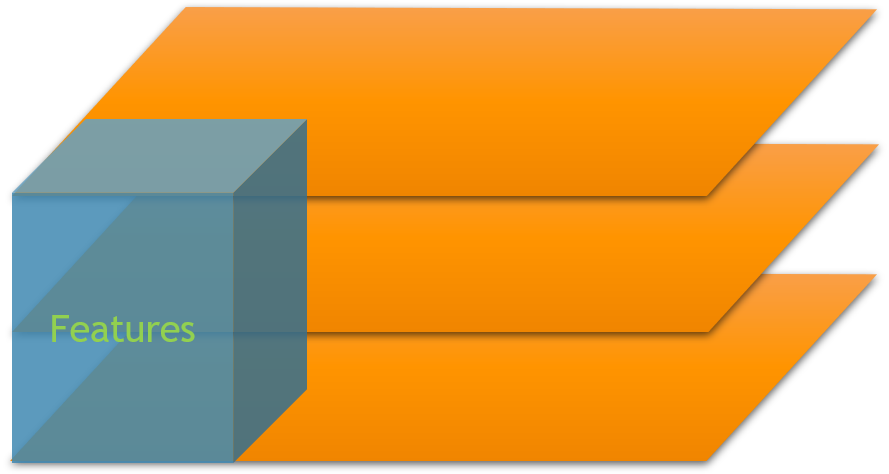}
		\caption{3D two stage segmentation - first stage}
		\label{fig:graphical-e}
	\end{subfigure}
	\begin{subfigure}{.3\textwidth}
		\includegraphics[width=\linewidth]{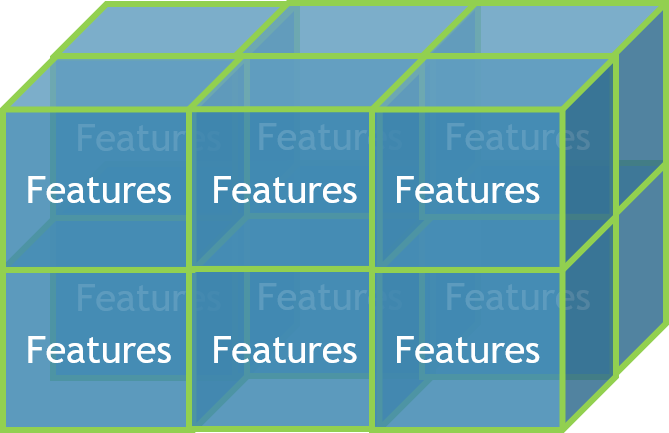}
		\caption{3D two stage segmentation - second stage}
		\label{fig:graphical-f}
	\end{subfigure}
	\caption{Graphical representation of the used input and output for the different proposed methods - 
		The orange slices represent the SHG images which form a scan together.
		The blue tiles or blocks are the inputs to the network while the green markings show the output format.
		For example (a) shows a 2D image part as an input and one value as output.
		While (f) takes features as a 3D matrix and outputs a 3D matrix.
	}
	\label{fig:graphical}
\end{figure}

\subsubsection{Classification}

A straight forward approach for local region classification is to create such regions by splitting the images into smaller image parts each with the same width and height.
An equal class distribution was enforced on these image regions and a graphical representation of the input is given in \autoref{fig:graphical-a}.
A rough segmentation for the image can be generated out of the classifications for all image parts.

\br
During development we discovered that  classification accuracy was higher if the image part size was larger.
These larger sizes lead to a rougher segmentation and thus we used an ensemble and majority voting to combine the benefits of a smaller image part size and the higher accuracy of larger image parts.
We used an Inception-ResNet-v2 \cite{incepresv2} backbone with pretrained weights, $\gamma = 0$ and $w = \{1\}^3$.

\subsubsection{Semantic segmentation}

Classification of image parts has two major drawbacks.
It is time consuming since a lot of image parts have to be processed and a post processing step is needed to combine the classifications to a segmentation.
A parallel rough semantic segmentation of an image can overcome both these problems.
In contrast to other literature \cite{fcn,unet} we use only a downscaling part and not an upsampling part in our segmentation network.
As described earlier we are not interested in a fine segmentation and can drop the upsampling because of this (see \autoref{fig:graphical-b}).

\begin{figure}[tb] 
	\centering
	\begin{subfigure}{.39\textwidth}
		\includegraphics[width=\linewidth]{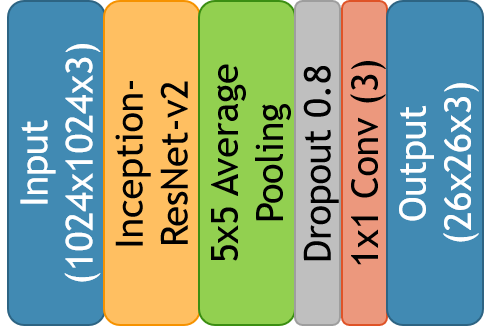}
		\caption{2D semantic segmentation}
		\label{fig:architecture-a}
	\end{subfigure}
	\begin{subfigure}{.55\textwidth}
		\includegraphics[width=\linewidth]{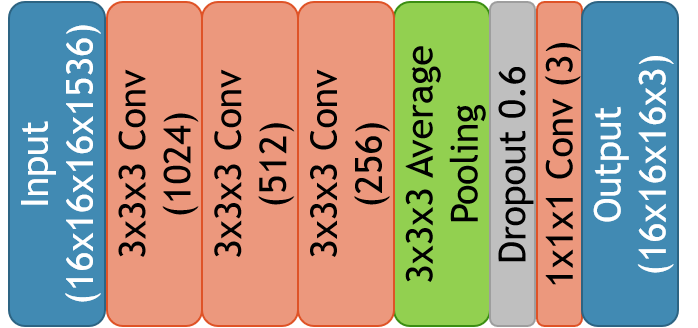}
		\caption{3D two stage segmentation}
		\label{fig:architecture-b}
	\end{subfigure}
	\caption{Architecture of the 2D semantic segmentation network with backbone and the second stage network for the 3D two stage segmentation - 
		Different layers have different colorings.	
		The main part of Inception-ResNet-v2 up to the global average pooling layer is described as one block.		
		The dimensions for in- and output and the number of feature maps for convolutional layers are given in brackets.
		The kernel size is given before the layer name.}
	\label{fig:architecture}
\end{figure}

\br
The segmentation networks differ from the original backbones mostly in the output.
The architecture is shown in \autoref{fig:architecture-a}.
We use an average pooling layer and a 1x1 convolutional layer instead of a global average pooling layer and a fully connected layer for multiple reasons.
Firstly, we want to create a matrix as an output which consists of softmax outputs for every row and column.
Secondly, we can incorporate the neighborhood information through the pooling layer.
Thirdly, we have to use a convolutional layer for the output because otherwise the number of parameters for the fully connected layer would have become unmanageable.
In addition we can create a finer segmentation than the ensemble above (\autoref{subsec:2d-methods}) and also use neighborhood information due to average pooling layers.
We used an Inception-ResNet-v2 \cite{incepresv2} backbone with pretrained weights, $\gamma = 0$ and $w = (\frac{16/}{41}, \frac{24}{41}, \frac{1}{41})$.
The weights are needed to account for the class imbalance in the input data.

\subsection{3D methods}
\label{subsec:3d-methods}

\subsubsection{Combination of 2D classifications}

This method is an extension of the 2D classification by combination.
We use the 2D classifications and include 3D information by averaging over the classifications which are positioned next to each other.
This simple aggregation yields a 3D classification but inherits the drawbacks of the 2D classification.
A graphical representation is shown in \autoref{fig:graphical-c}.

\subsubsection{Classification}

This method is a extension of the 2D classification in 3D.
Instead of image parts (square) we use scan blocks (cuboid) for the classification.
The graphical representation of the input and output is shown in \autoref{fig:graphical-d}.
However, we do not use an ensemble to combine different 3D classifications.
We used our proposed Inception-ResNet-3D with transferred 2D weights based on ImageNet, $\gamma = 2$,  $w = \{1\}^3$ and used the transfer strategy based on \autoref{eg:transfer1}.

\subsubsection{Two stage segmentation}

In the 2D case parallel segmentation of a complete image could utilize the neighborhood information for every entry in the output matrix.
In order to combine the information of a whole scan for an output and still fit in the memory of one GPU we had to take a two stage approach.
The idea is to extract features with a pretrained network and then combine these features in a second network to create a 3D matrix where every entry correspondence to the three classes (graphical representation see \autoref{fig:graphical-e} and \autoref{fig:graphical-f}).

\br
We used Inception-ResNet-3D as an extraction network with transferred ImageNet weights with the transfer strategy based on \autoref{eg:transfer1}.
Unlike in 3D classification we do not want one classification but the features as output.
The second stage is a small network out of convolutional and average pooling layer to combine the 3D matrix of features to class predictions. 
The architecture is shown in \autoref{fig:architecture-b} and was inspired by \autoref{fig:architecture-a}.

\section{Experimental Results}
\label{sec:results}

\subsection{Dataset}
\label{subsec:data}

We developed our methods on one dataset which was created by the MOIN CC\footnote[1]{\textbf{Mo}lecular \textbf{I}maging \textbf{N}orth \textbf{C}ompetence \textbf{C}enter}.
The dataset consists of 4736 SHG images from 35 scans of 6 mice where 3 mice had the disease OI and the others do not.
The scans were taken on different parts of the legs and had a resolution of 1000 x 1000 px or 1024 x 1024 px while capturing 250  \si{\um} x 250  \si{\um} of the bone.
We cannot downscale these images because we would loose the necessary resolution fine fiber structures.
The depth of each scan was variable and ranged from 78 to 214 images while the distance in the bone between each image is 0.5 \si{\um}.

\begin{figure}[tb] 
	\centering
	\begin{subfigure}{.3\textwidth}
		\includegraphics[width=\linewidth]{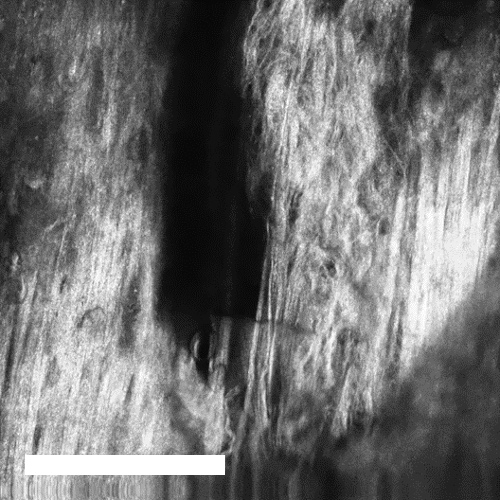}
		\caption{}
	\end{subfigure}
	\begin{subfigure}{.3\textwidth}
		\includegraphics[width=\linewidth]{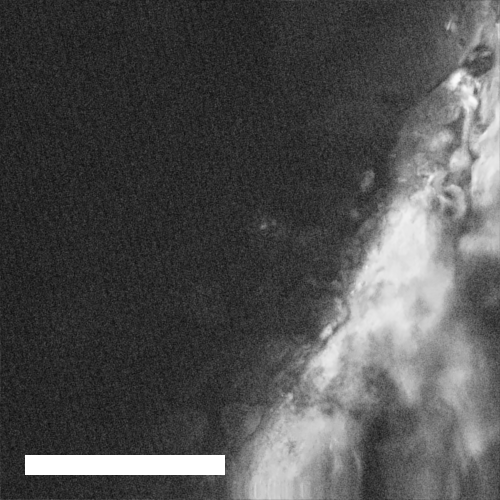}
		\caption{}
	\end{subfigure}
	\begin{subfigure}{.3\textwidth}
		\includegraphics[width=\linewidth]{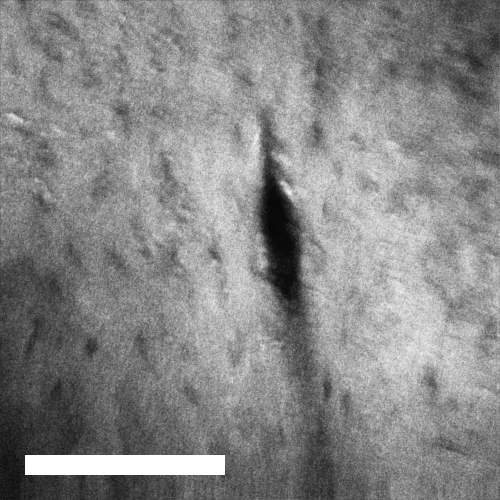}
		\caption{}
	\end{subfigure}
	\caption{Examples of used SHG images - 
		(a) shows a desirable image. We have  sharp and fine collagen structures while the noise ratio is low.
		(b) shows a noisy image where we can see collagen structures on the right hand side. Due to blurring and noise we can only detect rough shapes.
		(c) shows the macroscopic structure of the bone. We can detect the hole in the center but no single fibers or fiber bundles.
		The white scale bar depicts a size of 100 \si{\um}.}
	\label{fig:data-diversity}
\end{figure}

\br
A main property of the data is the class imbalance.
Roughly 2\% of the data belongs to the class S (similar orientation) and 3\% to the class D (dissimilar orientation).
The remaining 95\% belong to the class N (not of interest) and are, therefore, not interesting in medical research.
The data was split in to a training, validation and test set.
\autoref{fig:data-diversity} displays three example of used SHG images which represent the variety in the input data.

\br
Moreover, investigations of selected background regions showed a high scanner noise.
The average grey value (0-255) of the background should be zero but varied between 2.91 and 40.2.
On average the registered grey values differ from the real values with a mean of 9.18 and a standard deviation of 8.79. 

\subsection{Evaluation Metrics}
\label{subsec:metric}

We use an adapted accuracy function to measure the performance of our results.
In short we use the mean of accuracies per class as our accuracy measure.
Our function $meanacc$ for prediction $\hat{y} \in [n]^k$ and ground-truth $y  \in [n]^k$ with $n \in \mathbb{N}$ number of classes and $k \in \mathbb{N}$ entries is defined as follows

\begin{equation}
\label{eq:mean-acc}
meanacc(\hat{y},y) = \frac{1}{n} \sum_{j=1}^{n} 
\frac{\sum_{i=1}^{k} \mathbbm{1}_{\hat{y_i} = y_i, \hat{y_i} = j}}
{\sum_{i=1}^{k} \mathbbm{1}_{y_i = j}}.
\end{equation}
The function $meanacc$ has the benefit of being stable against class imbalance and is the same as the normal accuracy in the case of class balance.
It allows an estimation of performance in a single value without tuning weights.
In this paper accuracy on our data always refers to $meanacc$.

\subsection{Method comparison}

We used a strict data separation during development to be able to compare methods.
All hyperparameters were selected based on the validation set.
The test set was only used during method comparison.
In general we noticed two trends.
Firstly, better network performance on ImageNet translates to improved accuracies in all our methods.
This isn't noteworthy for tasks like 2D classification but for improvements in segmentation and feature extraction tasks it is.
Secondly, pretrained weights ensure a good initialization and lead to greater accuracies.
This is expected and reported for 2D classification tasks but the fact that pretraining can be interpolated to a 3D case and still ensures greater performance is significant.

\begin{table}[tb]
	\centering
	\begin{tabular}{l c c c c} \hline \addlinespace
		Method  &  Run-time & Resolution & Accuracy \\ \hline \addlinespace 
		2D Fourier analysis & N/A* & 16 x 16 x 1 & 55.00\%* \\ \addlinespace 
		2D classification  & 101 min &  64 x 64 x 1 &  59.54\%\\ \addlinespace
		3D combination of 2D classifications & 101 min &  64 x 64 x 16 & 59.68\%  \\ \addlinespace
		2D semantic segmentation  & \textbf{14 min} &  \textbf{40 x 40 x 1} & 64.58\% \\ \addlinespace
		3D classification  & 17 min & 128 x 128 x 64  & 70.51\%\\ \addlinespace
		3D two stage segmentation  & 72 min & 64 x 64 x 16 & \textbf{72.18\%}\\ \addlinespace	
	\end{tabular} 
	\caption{Overview of the best results on the test set for each method -
		Run-time is the time it took to process the test set once which includes pre- and postprocessing.
		Resolution describes the number of pixels in the input that are mapped to one output value. 
		Smaller resolutions result in finer segmentation but also in an accuracy drop for some methods and are, therefore, not reported here.
		The accuracy is reported on the test set.
		The best result of each column is marked bold.
		*Due to the inferior performance on the validation data and long run-time we did not evaluate the method Fourier analysis on the test set.
	}
	\label{tbl:results}
\end{table}

\br
\autoref{tbl:results} compares all presented methods with regard to run-time, resolution and accuracy.
The Fourier analysis results in the worst performance even on the validation set. 
Due to this inferior performance and the long run-time we did not evaluate the method further.
We believe that the high variability in the data can not be captured from such a simple approach.
The method of 2D semantic segmentation has the fastest run-time and the finest resolution. 
The accuracy is with about 65\% the best for all 2D methods.
The methods 3D classification and 3D two stage segmentation score a higher accuracy but have a rougher resolution and a longer run-time.
The best accuracy of 72\% is achieved by 3D two stage segmentation.

\br
In general we see that it is beneficial to process as much data as possible simultaneously to achieve a high accuracy.
This result can be explained due to the fact that simultaneous processing incorporates a larger neighborhood.
Furthermore, we see that there is not one best method in all regards.
Only a trade-off between different characteristics can be chosen.

\br
It is remarkable that the features used in the second stage were extracted with transferred 2D ImageNet weights and still lead to the best results.
Furthermore no adaption to domain specific weights is needed.
In the context of a human consistency of about 78.29\% in the user study (\autoref{sec:pre}) a result of 72\% is remarkable.

\subsection{Cross-validation}

We did 10 fold cross-validation on the complete dataset to verify that the chosen random split in different sets introduced no bias and represent the real data distribution.
The data was split 10 times into a training (50\%), a validation (25\%) and a test (25\%) set.
We split randomly but kept only splits where at least 2\% of the data had the class S and another 2\% the class D.
We put scans from the same bone region into the same set.

\br
We trained the method two stage segmentation on the training set, used the best weights on the validation set and evaluated on the test set.
The mean accuracy over 10 runs is 75.79\%.
\autoref{fig:cross-validation} shows that the accuracies for all runs are in a margin of about 10\% around the mean.
Some runs achieve an accuracy above the expected accuracy based on the user study.

\pic{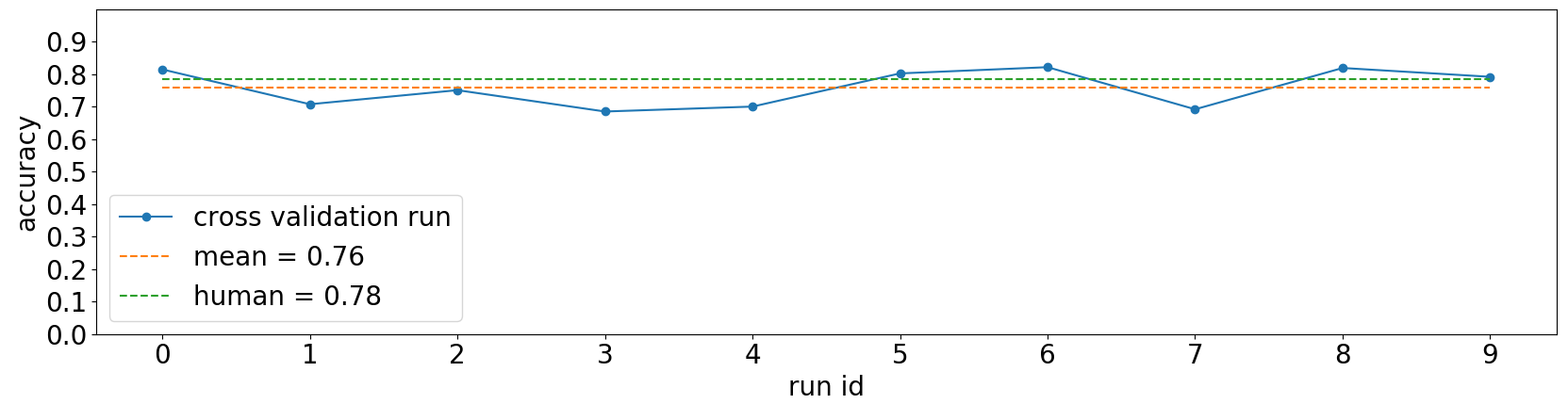}{Results of the cross-validation with mean and human performance based on the user study}

\section{Conclusion}

We compared a variety of methods for rough semantic segmentation of collagen fiber orientation in 2D and 3D.
As a dataset we used a novel collection of dense 3D SHG scans which is larger and more diverse as previously used datasets \cite{pork,ft-shg-3d}.
Our conducted user study implies that human can reach an average consistency of about 78.29\% on the task of uncertain collagen fiber orientation segmentation.
This results in a similar expected accuracy for trained algorithm due to the human annotated ground-truth.
We showed how to use transformed 2D ImageNet weights in 3D networks in general and in Inception-ResNet-3D in particular.
We proposed a two stage model that can simultaneously process large 3D inputs and use transformed 2D weights.
This best method two stage segmentation achieves an average accuracy of 75.79\% over 10 fold  cross-validation.
Based on the user study we can say that we created an algorithm with near human performance.

\br
The presented user study led to great insights into possible performance of neural networks.
It will be beneficial to repeat the user study at a larger scale.
We are confident that two stage segmentation with transferred weights can be applied in different 3D classification and rough segmentation tasks.
We will investigate these usages in the future.
Furthermore, we will investigate how to create more objective ground-truth for example by leveraging pretrained features and reduced supervision.

%
% ---- Bibliography ----
%
\newpage
\bibliographystyle{splncs04}
\bibliography{references}

\end{document}